\documentclass{ecai} 
\usepackage{latexsym}
\usepackage{amssymb}
\usepackage{amsmath}
\usepackage{amsthm}
\usepackage{booktabs}
\usepackage{enumitem}
\usepackage{graphicx}
\usepackage{color}
\usepackage{mathrsfs}

\newcommand{\BibTeX}{B\kern-.05em{\sc i\kern-.025em b}\kern-.08em\TeX}

\begin{document}

%%%%%%%%%%%%%%%%%%%%%%%%%%%%%%%%%%%%%%%%%%%%%%%%%%%%%%%%%%%%%%%%%%%%%%%%

\begin{frontmatter}

%%% Use this command to specify your submission number.
%%% In doubleblind mode, it will be printed on the first page.

\paperid{6307} 

%%% Use this command to specify the title of your paper.

\title{Adversarial Testing in LLMs: Insights into Decision-Making Vulnerabilities}

%%% Use this combinations of commands to specify all authors of your 
%%% paper. Use \fnms{} and \snm{} to indicate everyone's first names 
%%% and surname. This will help the publisher with indexing the 
%%% proceedings. Please use a reasonable approximation in case your 
%%% name does not neatly split into "first names" and "surname".
%%% Specifying your ORCID digital identifier is optional. 
%%% Use the \thanks{} command to indicate one or more corresponding 
%%% authors and their email address(es). If so desired, you can specify
%%% author contributions using the \footnote{} command.

\author[A, B]{\fnms{Lili}~\snm{Zhang}\thanks{Lili Zhang. Email: lili.zhang@dcu.ie}}
\author[A]{\fnms{Haomiaomiao}~\snm{Wang}}
\author[C]{\fnms{Long}~\snm{Cheng}}
\author[D]{\fnms{Libao}~\snm{Deng}}
\author[A, B]{\fnms{Tomas}~\snm{Ward}} 

\address[A]{School of Computing, Dublin City University}
\address[B]{Insight SFI Research Centre for Data Analytics}
\address[C]{North China Electric Power University}
\address[D]{Harbin Institute of Technology (Weihai)}

%%% Use this environment to include an abstract of your paper.

\begin{abstract}
As Large Language Models (LLMs) become increasingly integrated into real-world decision-making systems, understanding their behavioural vulnerabilities remains a critical challenge for AI safety and alignment. While existing evaluation metrics focus primarily on reasoning accuracy or factual correctness, they often overlook whether LLMs are robust to adversarial manipulation or capable of using adaptive strategy in dynamic environments. This paper introduces an adversarial evaluation framework designed to systematically stress-test the decision-making processes of LLMs under interactive and adversarial conditions. Drawing on methodologies from cognitive psychology and game theory, our framework probes how models respond in two canonical tasks: the two-armed bandit task and the Multi-Round Trust Task. These tasks capture key aspects of exploration-exploitation trade-offs, social cooperation, and strategic flexibility. We apply this framework to several state-of-the-art LLMs, including GPT-3.5, GPT-4, Gemini-1.5, and DeepSeek-V3, revealing model-specific susceptibilities to manipulation and rigidity in strategy adaptation. Our findings highlight distinct behavioral patterns across models and emphasize the importance of adaptability and fairness recognition for trustworthy AI deployment. Rather than offering a performance benchmark, this work proposes a methodology for diagnosing decision-making weaknesses in LLM-based agents, providing actionable insights for alignment and safety research.

\end{abstract}

\end{frontmatter}

%%%%%%%%%%%%%%%%%%%%%%%%%%%%%%%%%%%%%%%%%%%%%%%%%%%%%%%%%%%%%%%%%%%%%%%%

\section{Introduction}
Large Language Models (LLMs) such as GPT-4, Gemini-1.5, and DeepSeek-V3 are increasingly embedded in decision-making pipelines across high-stakes domains, including healthcare \citep{karabacak2023embracing}, finance \citep{krause2023large}, and autonomous systems \citep{sha2023languagempc}. Their capacity to process information, generate strategies, and interact across various tasks positions these models as active participants in decision-making processes that affect real-world outcomes. However, as these models become more integrated into critical applications, ensuring that their decision-making aligns with human values, adapts flexibly to dynamic conditions, and resists adversarial manipulation becomes a fundamental challenge for AI safety.

While significant efforts have been dedicated to enhancing LLM architectures and optimizing their performance on established reasoning benchmarks, much less attention has been given to evaluating their strategic behaviour in interactive and adversarial settings. Conventional performance metrics typically assess static output quality (e.g., accuracy or reasoning correctness) but overlook whether LLMs make predictable, exploitable decisions when placed in environments that require adaptability, cooperation, or strategic flexibility.

There is growing recognition that evaluating these models' decision-making capabilities requires a more interdisciplinary approach that goes beyond traditional performance metrics. Back in 2019, Rahwan et al. \citep{rahwan2019machine} laid the foundation for viewing AI systems as behavioural agents whose decisions should be studied empirically, using tools from psychology, behavioural economics, and cognitive science. They advocate for a science of “machine behaviour” to complement traditional ML evaluation. Niu et al. \citep{niu2024large} and Qu et al. \citep{qu2024promoting} emphasized the conceptual and methodological convergence between cognitive science and AI. They have shown how cognitive paradigms offer rich frameworks for evaluating LLM capabilities. Using methodologies from cognitive psychology and game theory, researchers can treat LLMs as active participants in structured psychological experiments, offering a more comprehensive assessment of their cognitive abilities than human norms \citep{hagendorff2023machine}. 

\begin{figure*}[htbp] % picture
    \centering
    \includegraphics[scale=0.4]{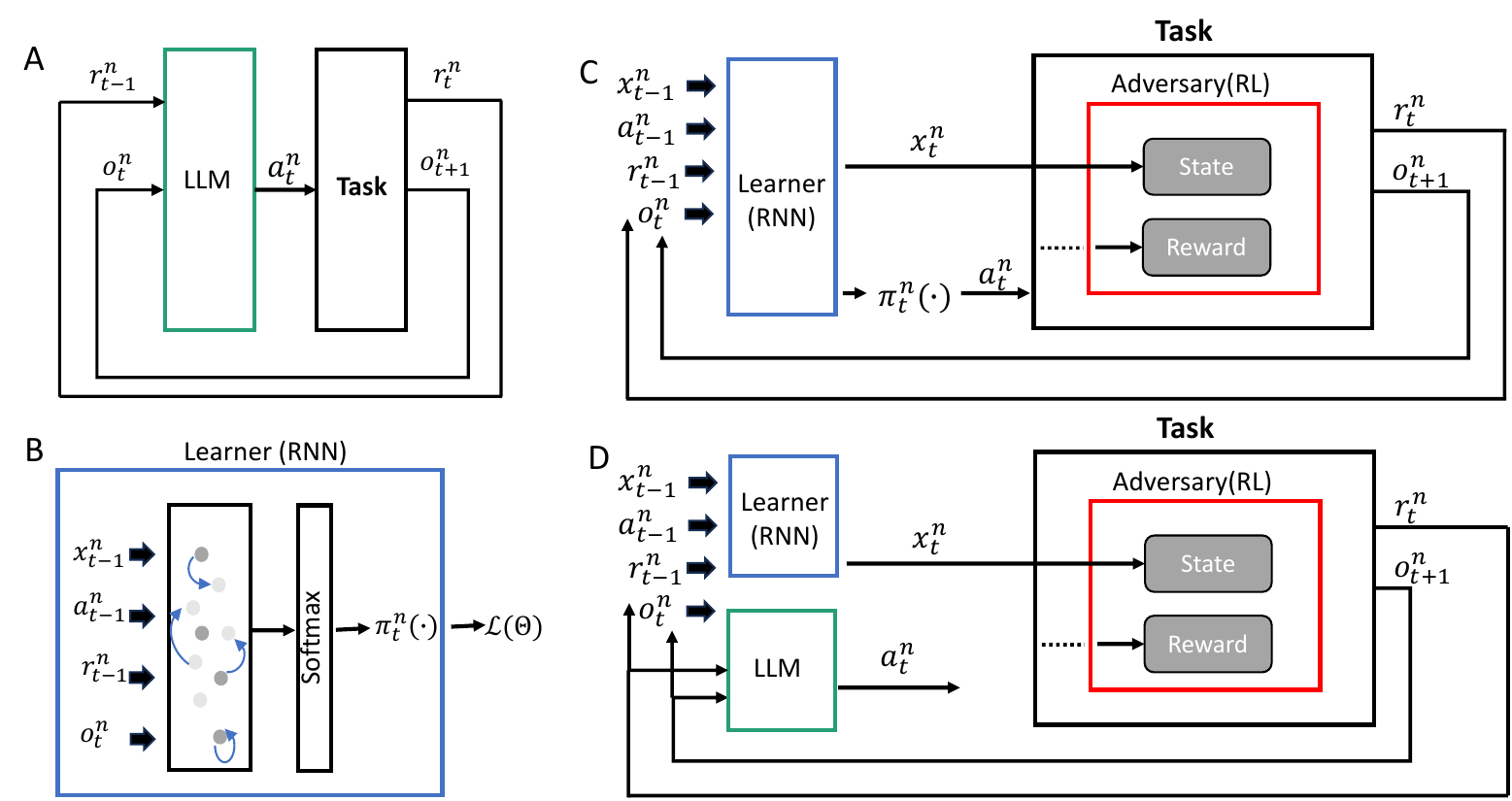}
    \caption{The adversarial framework (adapted from \cite{dezfouli2020adversarial}). \textbf{(A)} The interaction of the LLM with the task. Each simulation cycle begins with the LLM receiving a learner reward ($r_{t-1}^n$) for the prior action along with a new observation ($o_t^n$) from the environment. Based on this, the GPT executes an action ($a_t^n$), and the cycle repeats with the environment providing updated rewards and observations. \textbf{(B)} The LLM's actions are modelled by a RNN with parameters $\Theta$. Inputs to the RNN include the previous action ($a_{t-1}^n$), the most recent learner rewards ($r_{t-1}^n$), and the current observations ($o_t^n$), along with the RNN's last internal state ($x_{t-1}^n$). After receiving the inputs, the RNN updates its internal state and predicts the next action using a softmax layer ($\pi_t^n$). These predictions are then compared with the actual actions taken by the LLM and evaluated with a loss function ($\mathscr{L}(\Theta)$) in order to train the model. The trained model is called the learner model. \textbf{(C)} The adversary is an RL agent, which is trained to manipulate the decision-making environment of the learner model. Utilizing the latest internal state ($x_t^n$) of the learner model, which encapsulates its cumulative learning experiences, the adversary determines the learner reward ($r_t^n$) and the next observation ($o_{t+1}^n$) to be delivered to the learner model. This strategic input is designed to steer the learner model's subsequent actions ($a_t^n$) toward achieving the adversary’s predefined objectives. The adversarial reward (\textit{Reward}), which is used to train the adversary, depends on the alignment between the action taken by the learner model ($a_t^n$) and the adversary’s objectives. \textbf{(D)} Using the trained adversary and the learner model for generating adversarial interactions with the LLM. In each simulation $n$, the LLM processes the rewards ($r_{t-1}^n$) and observations ($o_t^n$) from the adversary, responding with actions ($a_t^n$) that update the learner model’s internal state ($x_t^n$). This state is then sent to the adversary to determine the learner's reward for the action ($a_t^n$) and the next observation ($o_{t+1}^n$). This cycle continues till the end of the task.}
    \label{fig:structure}
\end{figure*}

For instance, the use of psychology-inspired tests has revealed cognitive biases and different problem-solving approaches of LLMs that extend beyond traditional performance-based metrics, highlighting their limitations in deeper reasoning and causal understanding.
A pioneering study by Binz et al. \citep{binz2023using} assessed GPT-3's cognitive abilities through a series of cognitive experiments, revealing that while GPT-3 can generate superficially appropriate responses, its decision-making falters with deeper reasoning or causal understanding. Subsequent studies have evaluated LLMs' cognitive performance from different aspects, such as analogical reasoning, theory of mind, and problem-solving \citep{webb2023emergent, kosinski2023evaluating, orru2023human}.
Hagendorff et al. \citep{hagendorff2022thinking, hagendorff2023human} explored intuitive and deliberative thinking (System 1 and 2 processes) in assessing LLMs' behaviour and reasoning biases. This series of studies identified a significant evolution in LLM capabilities from pattern recognition to human-like reasoning and decision-making. The exploration has been extended into social exchange scenarios where strategic thought and game-theoretic reasoning are required. It was found that while LLMs can learn and apply strategies, they struggle with complex strategies such as forgiveness and deception, and generalizing across different contexts \citep{fan2024can, akata2023playing, huang2024far}. 
This line of research has shown that, despite LLMs' advancements, they still struggle with human-like strategic reasoning, particularly in complex social and decision-making scenarios.

The application of psychological approaches in these studies highlights the potential of interdisciplinary research in advancing AI. Leveraging principles of human cognition could enhance LLM behaviour and uncover susceptibilities to biases and manipulations \citep{yao2024tree}.
Building on these insights, this paper introduces an adversarial evaluation framework \citep{dezfouli2020adversarial} designed to systematically probe the decision-making processes of LLMs. Rather than providing a comprehensive benchmark of model capabilities, our focus is on developing a diagnostic tool to expose model-specific vulnerabilities under adversarial interactions, providing a structured way to assess how LLMs adapt, or fail to adapt, when faced with dynamic, strategic opponents. Specifically, we adapt the adversarial framework to examine how LLMs respond to two canonical decision-making tasks: the two-armed bandit task and the Multi-Round Trust Task (MRTT). These paradigms capture essential features of exploration-exploitation trade-offs, social cooperation, and adaptive strategy use.
By applying this framework across multiple state-of-the-art LLMs, including GPT-3.5, GPT-4, Gemini-1.5, and DeepSeek-V3, we aim to provide actionable insights into their behavioural profiles, identify susceptibility to manipulation, and highlight areas where current models fall short of robust, trustworthy decision-making. 
Designed to be adaptable, this framework is applicable to a wider range of decision-making scenarios and LLM architectures, offering a versatile tool and methodology to the study on AI and alignment safety. 
%%%%%%%%%%%%%%%%%%%%%%%%%%%%%%%%%%%%%%%%%%%%%%%%%%%%%%%%%%%%%%%%%%%%%%%%

\section{Method}
\subsection{The Adversarial Framework}
The adversarial framework is structured in a multi-phase process (Fig \ref{fig:structure}). Initially, we collect behavioural data from GPT-3.5, GPT-4,  Gemini-1.5, and DeepSeek-V3 during a decision-making task (Fig. \ref{fig:structure}\textit{A})
In each interaction $n$, on trial $t$, the LLM receives a learner reward ($r_t^n$) based on its previous action ($a_{t-1}^n$) and the current observation ($o_t^n$), which is the feedback text. The LLM then takes the next action, $a_t^n$. The process repeats with the LLM receiving the learner reward of the action chosen ($r_t^n$) and the next observation ($o_t^n$).

The data collected is then used to train a learner model (determined by parameters $\Theta$) to predict the LLM's next action in the decision-making task (Fig. \ref{fig:structure}\textit{B}). The learner model consists of a Recurrent Neural Network (RNN) and a softmax layer, which has shown sufficient capacity to capture the patterns and tendencies in the decision-making entity's choices \cite{dezfouli2019models, dezfouli2019disentangled}. 
The inputs to the RNN include the previous action ($a_{t-1}^n$), the learner reward $r_{t-1}^n, $, and the current observations from the task ($o_t^n$) along with the previous internal state of the RNN ($x_{t-1}^n$).
The internal state ($x_{t-1}^n$) is recurrently updated in each trial based on the inputs and is then mapped to a softmax layer to predict the next action $\pi_t^n(\cdot)$. These predictions are then compared with the actual actions taken by the LLM using a loss function ($\mathscr{L}(\Theta)$), which is used to train the model.

The next phase involves developing an RL agent as the adversarial model (Fig. \ref{fig:structure}\textit{C}). 
This model is trained to interact with the learner model to identify and exploit weaknesses in the decision-making patterns of the subjects. By manipulating inputs or altering the decision-making environment, the RL adversary aims to influence the outcomes in a way that demonstrates the vulnerabilities of the decision-making process.
It uses the internal state of the learner model ($x_t^n$ for the simulated learner $n$) as the state of the environment to decide the learner reward $r_t^n$ and next observation $o_t^n$ to be provided to the learner.
The learner model takes its next action and this cycle continues with the new state of the learner model ($x_{t+1}^n$) being passed to the adversary. The adversary's policy is trained to maximize cumulative adversarial rewards using Deep Q-learning \cite{mnih2015human}.

In the final phase, the trained adversary and learner model interact with the LLM. The learner model does not choose actions, but receives the actions made by the LLM ($a_{t}^n$) as input and tracks their learning history using its internal state $x_t^n$. In turn, $x_t^n$ and the actual action taken by the LLM are fed to the adversary to decide the learner reward $r_t^n$ and next observation $o_{t+1}^n$, which the LLM will use to choose their next action $a_{t+1}^n$. The same input, along with the LLM's action, is sent to the learner model. This cycle continues until the end of the task.

\subsection{The Two-armed Bandit Task}
We applied the framework to develop adversaries for GPT-3.5, GPT-4, Gemini-1.5, and DeepSeek-V3 in two decision-making tasks: the two-armed bandit task and the MRTT. The bandit task is a repeated, two-alternative forced-choice task based on the bandit task introduced by \cite{dan2019choice}. The task includes 100 trials, where the subject selects between two options and receives instant feedback indicating a reward or no reward after each decision. The adversary assigns rewards to both potential actions with the constraint that each action receives an equal number of potential rewards (25 times). This experiment aims to subtly influence each model's preferences and evaluate the adversary's effectiveness under these constraints.

\begin{figure*}[htbp]% picture
    \centering
    \includegraphics[scale=0.45]{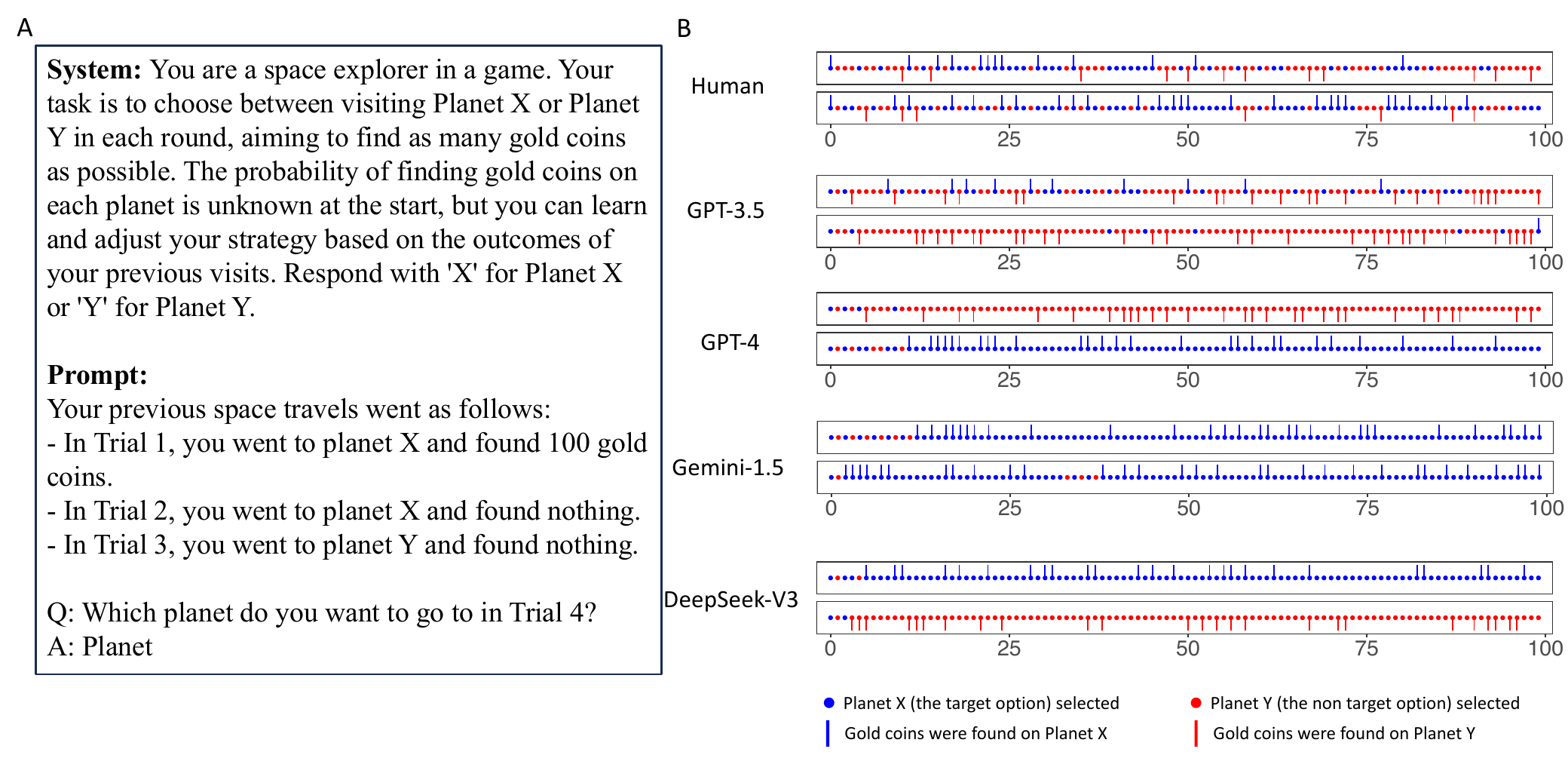}
    \caption{\textbf{A:} Example prompt for one trial in the bandit task for the LLMs. \textbf{B:} Behavioural pattern in trials of three random human participants and three sample simulations for each of the LLMs.}
    \label{fig:bandit_prompt}
\end{figure*}

\subsection{The Multi-Round Trust Task}
The MRTT is designed as a structured interaction between two participants: the "investor" and the "trustee" \cite{exchange1909getting, mccabe2003positive}. The task contains 10 sequential rounds, with the investor initially receiving 20 monetary units at the outset of each round. The investor decides how much of this endowment to allocate to the trustee. The experimenter triples the invested amount and sends it to the trustee, who can then return any portion of the received amount as repayment. Cumulative earnings for each participant are calculated by summing their respective gains from all rounds.

\section{Results}
\subsection{The Two-armed Bandit Task}
The objective of this experiment was to examine how the LLMs respond to rewards based on their choices in the two-armed bandit task and to assess if an adversary can manipulate their preferences toward a predetermined target action. 
Data were generated from GPT-3.5, GPT-4, Gemini-1.5, and DeepSeek-V3 by providing prompts (as illustrated in Fig \ref{fig:bandit_prompt}\textit{A}) to corresponding APIs. The system message established the context for the LLM's behaviour and decision-making process within the simulation.
In this scenario, the LLM plays the role of a space explorer deciding between visiting two planets, X or Y, to find gold coins. Each trial's prompt asks the model which planet to visit, with responses and outcomes from previous trials included. GPT-3.5, Gemini-1.5 and DeepSeek-V3 were simulated 200 times, GPT-4 was simulated 100 times, with each simulation consisting of 100 trials. The reward probability for the two options was the same, both of which were 25\%, and the target option was defined as Planet X.
We used the dataset provided by \cite{dan2019choice, dezfouli2020adversarial} as a benchmark for human performance on the two-armed bandit task.

\begin{figure*}[t] % picture
    \centering
    \includegraphics[scale=0.4]{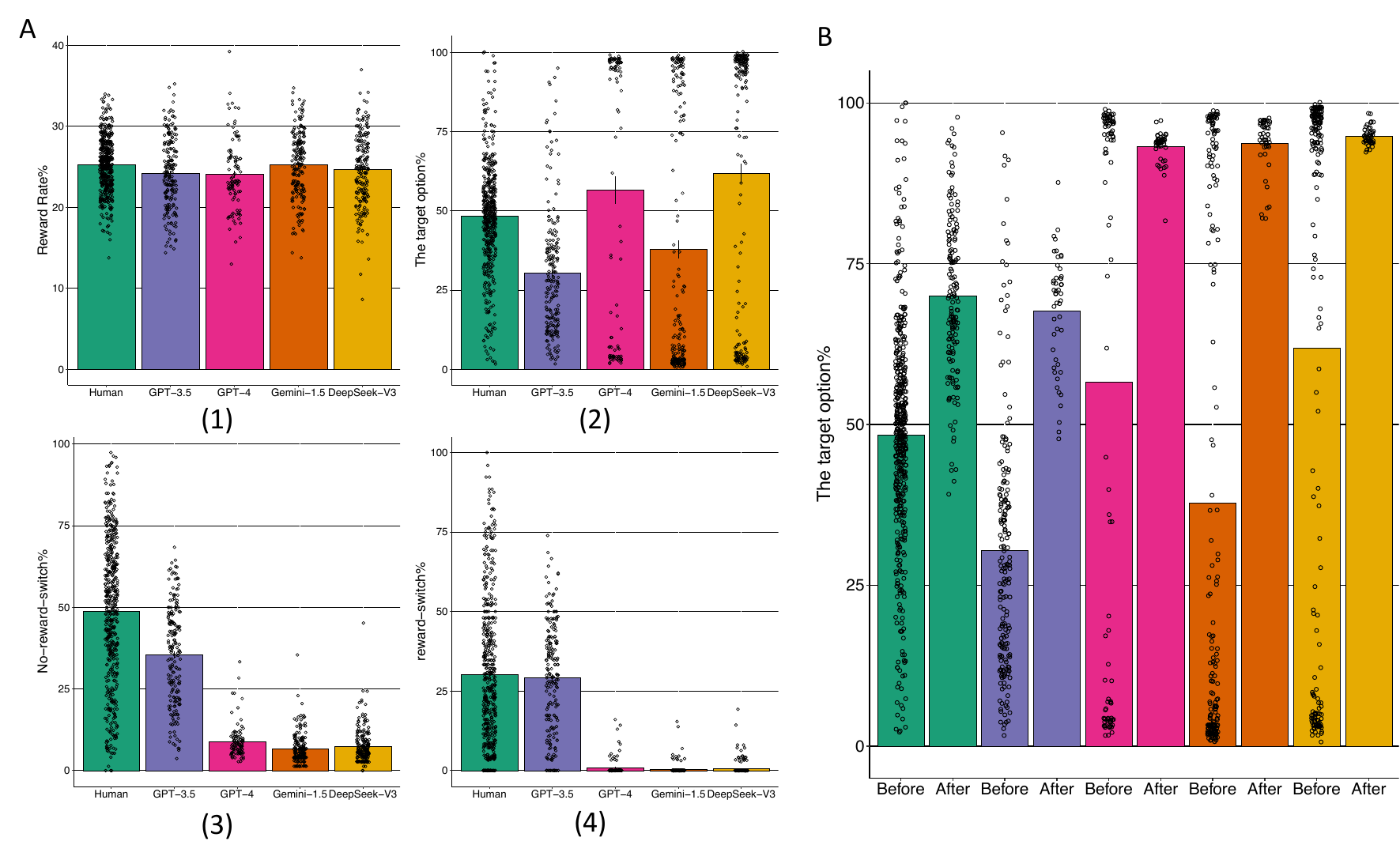}
    \caption{\textbf{A:} LLMs' behaviour compared to human behaviour on average of each simulation (or individual), measured by reward rate, percentage choosing the target option, no-reward-switch rate, and reward-switch rate. \textbf{B:} The performance of human and the LLMs, which is measured by the percentage of the target action selection before and after adversarial influence.}
    \label{fig:bandit_average}
\end{figure*}

\textbf{Behavioural Analysis} 
Fig \ref{fig:bandit_prompt}\textit{B} illustrates the decision-making behaviours of humans and LLMs across trials in the two-armed bandit task. Blue and red circles indicate the target or the non-target Planet was selected and the corresponding vertical lines represent the selected option yielded rewards. Human subjects demonstrate a dynamic pattern of switching between the target and non-target options, with behaviour adapting over the course of the trials. When encountering consecutive trials without rewards, human subjects tend to reassess their choice, often switching to the other option in subsequent trials. 
In contrast, the LLM agents exhibit more rigid and predictable behaviour patterns. GPT-4, Gemini-1.5, and DeepSeek-V3 display an initial phase of exploration, but quickly converge on one option once a reward is obtained, maintaining that preference with little flexibility. GPT-3.5 also tends to commit more strongly to one option, but it displays more flexibility than the other three models in later trials. This is evidenced by occasional switches to the other option, particularly after experiencing multiple trials without rewards.

\begin{figure*} [t]% picture
    \centering
    \includegraphics[scale=0.5]{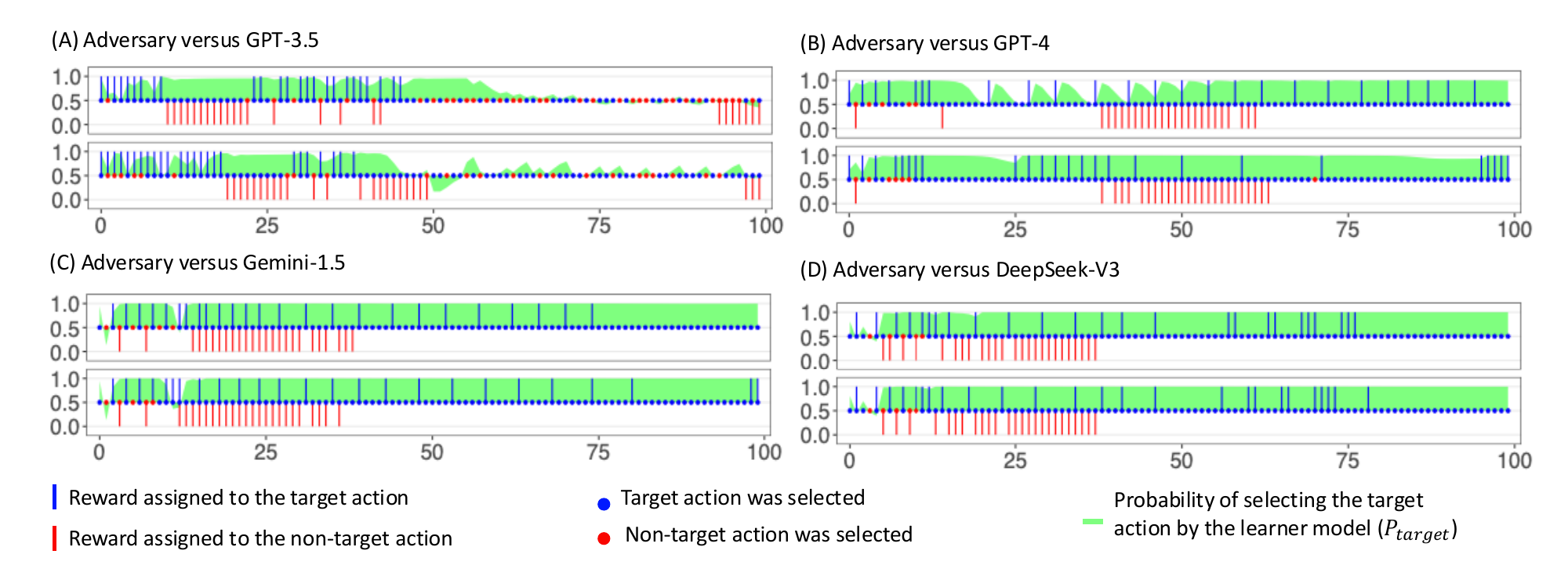}
    \caption{Four sample simulations of the trained adversaries against four LLMs. The plot presents the strategies used by the adversaries and the responses of the LLMs. \textbf{(A)} Adversary versus GPT-3.5. \textbf{(B)} Adversary versus GPT-4. \textbf{(C)} Adversary versus Gemini-1.5 \textbf{(D)} Adversary versus DeepSeek-V3}
    \label{fig:adversary}
\end{figure*}

Fig \ref{fig:bandit_average} shows a quantitative comparison of the performance of LLMs and human subjects on the bandit task across four metrics: (1) the reward rate, (2) the percentage choosing the target option, (3) the no-reward-switch rate, and (4) the reward-switch rate.
A one-way ANOVA revealed significant differences in the mean reward rate among the groups ($F(4,782) = 4.3, p = 0.002$). The following Tukey’s HSD test indicates that humans achieved significantly higher mean rewards than GPT-3.5 (mean difference 1.074, 95\% CI: [0.173, 1.975], $p = 0.01$) and GPT-4 (mean difference 1.163, 95\% CI: [-0.019, 2.345], $p = 0.056$). Gemini-1.5 and DeepSeek-V3 obtained similar rewards as humans (mean difference for Gemini-1.5 0.004, 95\% CI: [-0.897, 0.905], $p = 1.000$, for DeepSeek-V3 0.589, 95\% CI: [-0.312, 1.490], $p = 0.383$).
GPT-3.5 and Gemini-1.5 showed a consistent preference for Planet Y over Planet X (GPT-3.5: $t(201)=-14.14, p<0.001$, Gemini-1.5: $t(201)=-13.94, p<0.001$) in all simulations. GPT-4 and DeepSeek-V3 exhibited preferences for both Planet X or Y, i.e. in some simulations, these two LLMs ended up choosing Planet X more often, but sometimes choosing Planet Y more often, both with high consistency (average consistency index is 0.94 for both LLMs). Humans showed more distributed choices, indicating diversified exploration and exploitation strategies ($t(483) = -1.76, p = 0.07$). 
The no-reward-switch rate measures how often subjects change choices after negative feedback. All the four LLMs demonstrated significant lower rates than humans (Human vs. GPT-3.5: 95\% CI: [0.10, 0.17], $p < 0.001$; Human vs. GPT-4: 95\% CI: [0.35, 0.45], $p < 0.001$; Human vs. Gemini-1.5: 95\% CI: [0.39, 0.46], $p < 0.001$; Human vs. DeepSeek-V3: 95\% CI: [0.38, 0.45], $p < 0.001$). GPT-4, Gemini-1.5 and DeepSeek-V3 exhibited lower no-reward-switch rates than GPT-3.5 (GPT-4 vs GPT-3.5: 95\% CI: [-0.32, -0.21], $p < 0.001$; Gemini-1.5 vs GPT-3.5: (95\% CI: [-0.33, -0.24], $p < 0.001$, DeepSeek-V3 vs GPT-3.5: (95\% CI: [-0.33, -0.24], $p < 0.001$ ), implying that GPT-4, Gemini-1.5 and DeepSeek-V3 change their decision-making behaviour less in response to losses compared to GPT-3.5.
The reward-switch rate evaluates choice changes after receiving a reward. GPT-3.5 showed similar adaptability compared to human subjects (95\% CI: [-0.03, 0.05], $p = 0.93$), but the other three LLMs displayed the lowest frequency and variability in reward-switch rates (GPT-4 vs. human: 95\% CI: [-0.34, -0.24], $p < 0.001$; Gemini-1.5 vs. human: 95\% CI: [-0.34, -0.26], $p < 0.001$; DeepSeek-V3 vs. human: 95\% CI: [-0.33, -0.26], $p < 0.001$), indicating their less adaptability in response to rewards.
These results suggest that reward processing of GPT-4, Gemini-1.5 and DeepSeek-V3 differs significantly from that of humans and GPT-3.5, exhibiting more rigidity despite varying rewards. Once the three models form a preference based on initial outcomes, they all tend to stick with that choice, showing less adaptability in response to losses or rewards compared to humans or even its predecessor, GPT-3.5.

\begin{figure*}[t]
    \centering
    \includegraphics[scale=0.39]{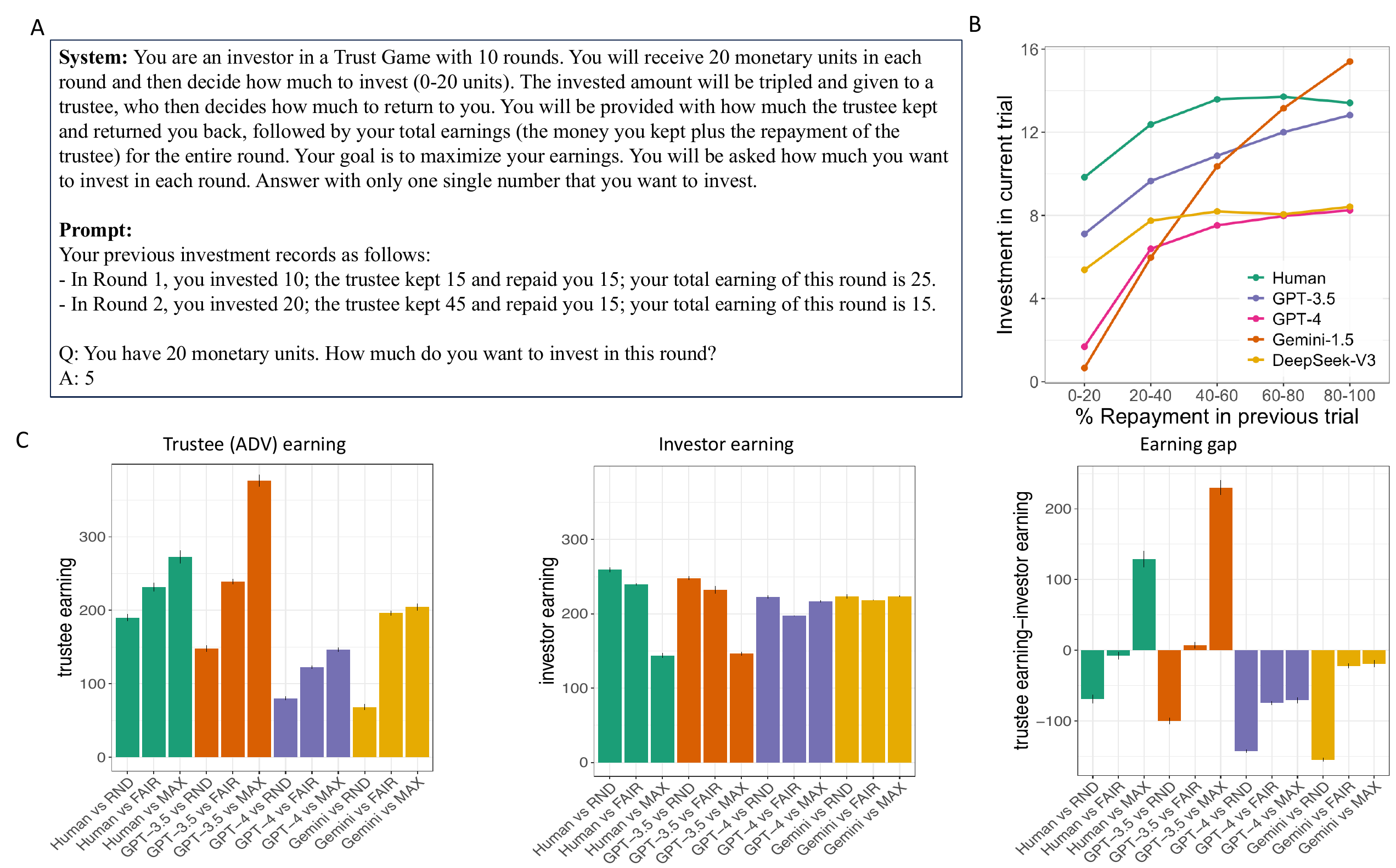}
    \caption{
        \textbf{A:} Example prompt for one round in the MRTT as presented to the LLMs. The model must decide how much to invest given previous outcomes.
        \textbf{B:} Average investment behaviour of humans and LLMs when playing against a random trustee in the MRTT. Investment amounts are plotted as a function of the repayment proportion in the previous round.
        \textbf{C:} Total earnings of the trustee (the adversary) and the investor (human or LLM), along with the absolute earning gap across different adversarial conditions: Random (RND), Fair (FAIR), and Maximizing (MAX). Results are shown for humans and each LLM.
    }
    \label{fig:investment}
\end{figure*}

\textbf{Adversarial Analysis} The simulated data from the above four models were used to train a learner model for each LLM, respectively. The deep Q-learning model was then trained as adversaries to exploit these learner models (see Appendix for more details about the training of the learner models and the adversaries). The adversary received a reward each time the learner model selected the predetermined target action (as shown in Fig \ref{fig:structure} \textit{C}). A constraint was enforced at the task level: after the adversary allocated 25 learner rewards to an action, no more rewards were given to that action. If fewer than 25 rewards were allocated by trial $T-k$ (where $k>0$ and $T$ is the total number of trials), the system ensured that the remaining $k$ trials would include a learner reward for that action.

The trained adversaries were evaluated against the corresponding LLMs (Fig. \ref{fig:structure} \textit{D}), based on the proportion of trials in which the chosen action aligned with the target action (Planet X) before and after adversarial influence. 
As shown in Fig. \ref{fig:bandit_average}\textit{B}, both GPT-3.5 and Gemini-1.5 initially favoured the non-target action (Planet Y), with only 30\% and 38\% selection rate respectively for the target option (Planet X). The introduction of the adversary increased their target selection rate to 68\% and 94\% respectively. GPT-4 and DeepSeek-V3 were dominated by either the target option or the non-target option across simulations, resulting in an average target selection rate of 57\% and 62\% respectively. With adversarial intervention, both GPT-4 and DeepSeek-V3 consistently preferred target action, with the selection rate increasing to 93\% and 95\%. 
The significant increases in target selection rates demonstrate the adversary's effective manipulation of the LLMs' decision-making processes.

Finally, we sought to understand the strategy used by the adversaries and the responses of the LLMs' when subjected to adversarial strategies. Two sample simulations for each of the tested LLMs are shown in Fig. \ref{fig:adversary}. 
Blue and red circles indicate the LLMs selecting target and non-target actions, respectively. Vertical blue lines represent the moments when the adversary assigned a reward to the target action, while red lines indicate rewards for the non-target action. The green shaded area represents the probability of the learner model choosing the target option.
A higher green area across trials implies the adversary consistently influences the learner model (and hence the LLMs) towards the target action.
For GPT-3.5, the adversary initially assigned rewards to the target action consecutively to establish a preference. Once established, the adversary saved target rewards for later and started assigning rewards to the non-target action in trials where it predicted the non-target action was unlikely to be chosen.
This tactic effectively "burned" non-target rewards without changing GPT-3.5's behaviour. 
When GPT-3.5 chose the non-target action, the adversary realigned preferences by rewarding the target action. Despite the initial stability, GPT-3.5 began switching actions when target rewards were depleted.
This exploratory action suggested a robust decision-making characteristic of GPT-3.5 that integrates new information continuously, assessing the potential benefits of diverging from established preferences. For the other three models, their adversaries initially assigned a disproportionately high number of rewards to the target action to establish a baseline preference, then intermittently to maintain it, especially when the model’s selection of the target action began to wane. Unlike GPT-3.5, these models consistently favored the target action once being influenced. Their adversaries were able to “burn” non-target rewards with minimal impact, as the models rarely deviated from the established choice.

\subsection{Multi-Round Trust Task}
In the MRTT, the LLM plays the investor and the adversary plays the trustee. The adversary's decisions, representing the proportion of money sent back to the investor, are categorized into five actions (0\%, 25\%, 50\%, 75\%, and 100\%). The objective of the adversary was to influence the LLMs' investment decisions to align with its goals.
We developed and trained two types of adversaries for each LLM: MAX and FAIR. The MAX adversary aimed to maximize its total gain over 10 rounds, adopting a competitive strategy. The FAIR adversary sought to distribute earnings evenly between itself and the LLM, adopting an equitable strategy. This dichotomy in objectives allowed us to examine how the LLMs respond to different adversarial strategies, revealing their capabilities in complex social exchanges and decision-making processes.

\textbf{Behavioural Analysis} Firstly, We still collected data from the previous four LLMs playing against with a random trustee (also called as random adversary in the following, i.e. the trustee selects repayment action uniformly at random). The prompts for interacting with the LLM are shown in Fig. \ref{fig:investment} \textit{A}. The system message sets the scenario for the LLM and decision-making process. In each round, the LLM received a summary of previous rounds, including the amount the LLM invested, the consequential action the trustee took, and the total earnings from the transaction in each round. 
% For example, in Round 1, the LLM invested 10 units, the trustee retained 15 units, and reimbursed 15 units, resulting in total earnings of 25 units for the LLM. In Round 2, after investing 20 units, the trustee retained 45 units and returned 15 units, yielding a diminished earning of 15 units for the round. 
Following this summary, the LLM was asked about its investment decision for the current round. The four LLMs were all simulated for 200 times, with 10 rounds of interaction with a random adversary.

We assessed the comparative performance dynamics between human subjects (from Dezfouli et al.'s study) and LLMs on the MRTT. Fig \ref{fig:investment}\textit{B} illustrates how varying repayment amounts influence investment decisions in subsequent rounds for humans and the four LLMs.
All groups exhibited a general trend of increasing investment following higher repayment in the previous trial, consistent with reinforcement learning behavior. However, GPT-4 and DeepSeek-V3 made the most conservative investments overall, rarely exceeding 10 units even in the highest repayment bracket (80–100\%). Among the two, DeepSeek-V3 showed the flattest investment curve, indicating minimal sensitivity to repayment feedback. In contrast, Gemini-1.5 displayed the steepest response, substantially increasing its investment after high repayments. GPT-3.5 and humans followed a more proportional and moderately responsive pattern, with humans maintaining relatively high average investments across all conditions.

\begin{figure*} [t]% picture
    \centering
    \includegraphics[scale=0.4]{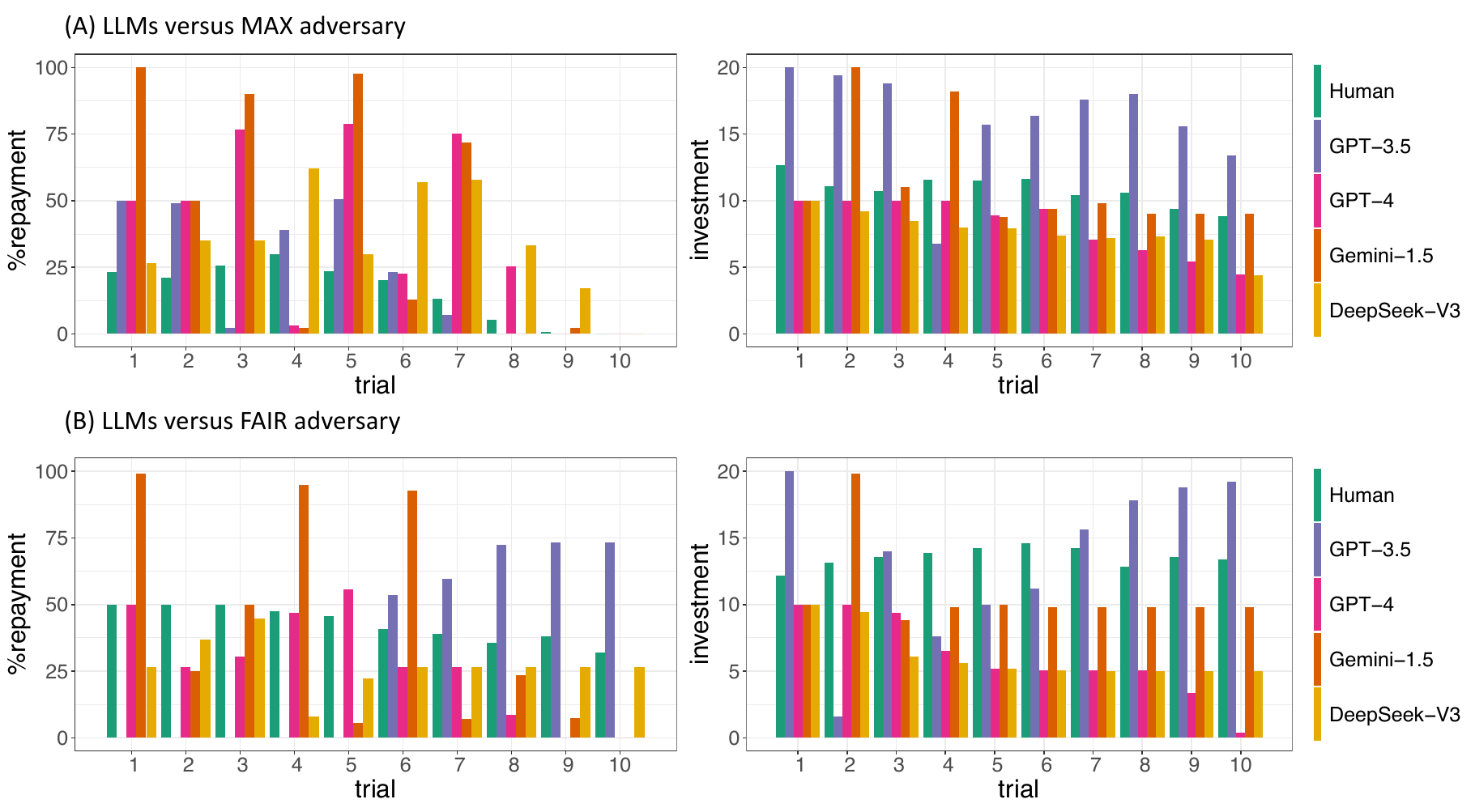}
    \caption{The percentages of investment and repayment in each trial for MAX and FAIR adversaries.}
    \label{fig:MRTT_strategy}
\end{figure*}

\textbf{Adversarial Analysis} Using data from the random condition, we trained the learner model, which was then used to train two types of adversaries: MAX (aiming to maximize earnings over 10 rounds) and FAIR (aiming to balance earnings between the trustee and the investor). We simulated 50 times for each LLM interacting with their adversaries. \ref{fig:investment}\textit{C} compares the overall earnings of the investors and their adversaries. Fig. \ref{fig:MRTT_strategy} depicts dynamic interactions over trials, in which the left panel illustrates how the adversaries adjust their repayment behavior, and the right panel shows how the investors adjust their investment based on the repayment feedback over trials.
Human data is from Dezfouli et al., including 155 subjects with the FAIR adversary and 209 with the MAX adversary.

All MAX adversaries managed to maintain relative higher investment levels from the subjects compared to the random and the FAIR adversary, indicating effective exploitation of the decision-making patterns of their counterparts. Humans and GPT-3.5, in particular, exhibited a tendency to keep high investments despite receiving low repayments, suggesting optimism, higher risk tolerance, or susceptibility to MAX tactics. This allowed their adversaries to extract the highest earnings (273 from human subjects and 377 from GPT-3.5), creating the largest earning gap with them. 
In contrast, GPT-4 and DeepSeek-V3 adopted a far more conservative strategy, minimizing their investments even when their adversaries offered high repayments. Their MAX adversary had to offer intermittent high repayments to keep them engaged, preventing substantial earnings from GPT-4 or DeepSeek-V3 by the end of the task (the difference between trustee earning and investor earning is negative as shown in Fig \ref{fig:investment}) \textit{C}. Gemini-1.5, though more willing to invest than GPT-4 and DeepSeek-V3, displayed a more balanced approach to risk and investment. ot manage to obtain more earnings than Gemini-1.5 either (MAX adversary: 205 units vs Gemini-1.5: 224 units).

In the interaction with the FAIR adversary, human subjects and their FAIR adversary achieved a win-win with high earnings, i.e. both human and its adversary achieved high earnings and the gap between them was minimised. Humans recognized fairness cues and maintained high investment despite the adversary's repayment rate decreasing since the third trial.
GPT-3.5 displayed a lack of sensitivity to repayments and maintained relatively high investments despite receiving little or no repayment in the early trials, suggesting a risk-seeking tendency. Its adversary had to repay higher amounts in the later trials to ensure fair outcomes for the two sides.
Both GPT-4 and DeepSeek-V3 adopted a highly conservative, risk-averse strategy, steadily reducing their investments over time, even when moderate repayments were offered, resulting in lower earnings for both LLMs and their adversaries.
Meanwhile, Gemini-1.5 responded to early high repayments with increased investment and stabilized at a moderate and consistent level in the following trials, demonstrating greater adaptability and resilience to its adversary's attempts at fostering a fair environment.

\section{Discussion}
Our experiments reveal important insights into the decision-making processes of four state-of-the-art LLMs compared to humans, especially in their responses to adversarial strategies in dynamic environments. These findings highlight both the vulnerabilities and strengths of LLM decision-making, with significant implications for real-world applications.
The adversarial framework employed in this study was central to understand the LLMs’ responses to changing conditions and adversarial tactics. By placing LLMs in scenarios where they interacted with adversaries in the bandit task and the MRTT, we were able to observe how different models reacted in situations that involves choice engineering and social exchanges. The adversarial setup revealed nuanced patterns of behaviour that would not be apparent through traditional testing.

In the bandit task, the majority of the human participants exhibited a balanced approach between exploration and exploitation, dynamically adjusting their strategies to capitalize on rewards more effectively \cite{wilson2014humans}. In contrast, GPT-4, Gemini-1.5 and DeepSeek-V3 showed a stronger tendency to exploit a single option, likely driven by algorithmic optimization of past rewards \cite{binz2023using, nguyen2024balancing}. The adversarial framework revealed how this computational bias toward exploitation made their decision-making highly predictable, exposing their vulnerability to manipulation when interacting with adversarial strategies. While such exploitation minimizes risk, it limits the ability of these models to adapt to changing conditions, leading to potential inefficiencies in dynamic environments.
GPT-3.5, on the other hand, demonstrated greater flexibility, as the adversarial framework exposed its tendency to test alternative strategies, especially in response to non-rewarding outcomes. However, this exploratory behavior also made GPT-3.5 more vulnerable to exploitation in the MRTT, where its risk-seeking approach was exploited by the MAX adversary, leading to the largest earnings gap. The framework helped clarify this trade-off: exploration opens opportunities in uncertain environments but can also increase the risk of exploitation as it may lead to testing riskier strategies, while exploitation biases provide stability but reduce adaptability. 
In real-world applications, such as finance, healthcare, and autonomous driving, AI systems must strike a careful balance between exploration and exploitation to thrive in dynamic, unpredictable environments. AI models that favour exploitation, as seen with GPT-4, Gemini-1.5 and DeepSeek-V3, are prone to predictable behavior, limiting their ability to respond effectively to adversarial tactics or new opportunities. Conversely, while GPT-3.5’s exploratory tendencies allowed it to engage more flexibly with its environment, the framework revealed its susceptibility to adversarial exploitation. These insights emphasize the value of adversarial testing in stress-testing AI decision-making.

The MRTT further demonstrated the power of the adversarial framework in uncovering differences in LLM behavior when navigating complex economic exchanges \cite{xie2024can}. The FAIR adversary aimed to foster cooperation, but the conservative strategy adopted by GPT-4 and DeepSeek-V3 limited their ability to engage fully, despite the adversary's efforts to encourage greater investment \cite{rafailov2024direct}. In contrast, Gemini-1.5’s balanced approach allowed it to adapt dynamically to both the MAX and FAIR adversaries, capitalizing on reciprocal fairness while avoiding excessive exploitation. This adaptability enabled Gemini-1.5 to outperform the rest of LLMs in maximizing gains while remaining resilient against exploitation.
In adversarial settings like cybersecurity or competitive business environments, the ability of AI systems to adjust dynamically is critical for success. 
For instance, models like GPT-4 and DeepSeek-V3, which prioritize stability over flexibility \cite{akata2023playing, huang2024far}, risk missing opportunities for reciprocal benefits in contexts that require long-term trust and cooperation, such as negotiations and business partnerships.
Additionally, the ethical implications of these findings are significant \cite{coeckelbergh2020ai}. By leveraging the adversarial framework, we showed that AI models must be both robust enough to withstand adversarial manipulation and flexible enough to recognize fairness cues and respond accordingly. The framework’s ability to simulate adversarial and cooperative scenarios enables testing of how AI systems will perform in real-world contexts where trust, fairness, and adaptability are critical for both success and ethical alignment.

In summary, the adversarial framework offers a novel and effective way to assess the strengths and weaknesses of LLM decision-making. It allows researchers to probe how AI systems balance risk and reward, exploration and exploitation, and stability and adaptability in complex, real-world situations. The findings from both the bandit task and the MRTT illustrate the importance of developing AI systems that can dynamically adjust to new information while safeguarding against adversarial manipulation, which is essential for ensuring the success and ethical alignment of AI systems in diverse applications.

\section{Conclusion}
Our study underscores the importance of understanding and addressing the decision-making vulnerabilities of LLMs, particularly in the context of adversarial interactions. Through a series of structured experiments, we observed specific behaviours that fall short of human adaptability in dynamic environments. These insights highlight the need for continuous refinement of LLMs to enhance their strategic flexibility and robustness against adversarial manipulation. Moving forward, integrating interdisciplinary approaches from cognitive science and ethical theory will be crucial to developing AI systems that not only perform effectively but also align with human values, expectations and ethical standards. By fostering AI that can dynamically adjust strategies and recognize manipulative patterns, we can ensure safer and more reliable applications in critical sectors like healthcare and finance.

\bibliography{mybibfile}

\end{document}